# Intelligent Bug Algorithm (IBA): A Novel Strategy to Navigate Mobile Robots Autonomously

Muhammad Zohaib[1], Syed Mustafa Pasha[2], Nadeem Javaid[3], Jamshed Iqbal[4]

[1,2,4] Department of Electrical Engineering, COMSATS Institute of Information Technology, Islamabad, Pakistan
[3] Center for Advanced Studies in Telecommunication (CAST), COMSATS Institute of Information Technology, Islamabad, Pakistan

[1]zohaib.ciit@yahoo.com, {[2]mustafa.pasha, [3]nadeemjavaid, [4]jamshed.iqbal}@comsats.edu.pk

**Abstract.** This research proposed an intelligent obstacle avoidance algorithm to navigate an autonomous mobile robot. The presented Intelligent Bug Algorithm (IBA) over performs and reaches the goal in relatively less time as compared to existing Bug algorithms. The improved algorithm offers a goal oriented strategy by following smooth and short trajectory. This has been achieved by continuously considering the goal position during obstacle avoidance. The proposed algorithm is computationally inexpensive and easy to tune. The paper also presents the performance comparison of IBA and reported Bug algorithms. Simulation results of robot navigation in an environment with obstacles demonstrate the performance of the improved algorithm.

## 1 Introduction

The revolution in the field of Mechatronics has made it possible to see the 'fiction' robots in reality in various fields of life ranging from articulated robots [1,2] to mobile robots [3]. Mobile robots permit human access to unreachable locations including accidental situations like fire, building collapse, fire, earthquake and hazardous scenerios such as Nuclear Power Plant (NPP) [4], chemical industry, transmission lines etc. Deployment of a mobile robot in real world applications demands addressing several new issues regarding robot interaction. The increasing development in robotics has brought up various challenges including obstacle avoidance, path planning, navigation, localization, autonomous control etc. Recent

This research has been funded by National Information and Communication Technology (ICT) R&D Fund under agreement no. NICTRRFD/ NGIRI/ 2012-13/corsp/3 Sr. 5.

Japanese robotic competitions on field robotics involving multi-disciplinary research highlighted challenging tasks [5].

Intelligence in the robot navigation to achieve autonomy is a challenging problem for researchers, as it is an important task to design the robot which can perform variety of tasks, such as surveillance, transportation, exploration or human locomotion [6,7]. The existing robot navigation algorithms can be categorized into three main types on the basis of knowledge of environment: completely known, partially known and unknown environment. In completely known environment, it is easy to tune a robot, by simply creating a map and applying A* search algorithm [8] to generate a reference path. On the other hand, in a partially or completely unknown environment, an obstacle avoidance algorithm is required. Obstacle avoidance is the backbone of autonomous control in the robot navigation especially in a fully unknown environment as it plays the key role in safe path planning. The algorithm for this purpose must be efficient enough, so that it can take a quick decision while encountering an obstacle without human intervention. Wide sample space of obstacle shapes that can be encountered in real world applications further necessitates the algorithm intelligence. In the field of mobile robotics, intelligent obstacle avoidance is the most important task, since every autonomous robot has to plan a safe path for its trajectory towards destination. This is achieved with an intelligent algorithm that uses knowledge of goal position and the sensorial information of the surrounding environment.

With a focus on these primary features, the present research proposes an intelligent goal-oriented algorithm for autonomous navigation of mobile robots. The proposed algorithm does outperform the existing approaches. It proves the convergence with relatively short, smooth and safe trajectoy.

The paper is organized as follows. Section 2 describes the related work. Section 3 introduces the proposed algorithm. Section 4 depicts the simulation results. Finally Section 5 comments on conclusion.

## 2   Related Work

Several algorithms are used for path planning with obstacle avoidance to navigate the mobile robots. The algorithm, which plans a shortest and smoothest path with obstacle avoidance capability, is considered as an ideal candidate for autonomous robots. Obstacle avoidance in some cases is difficult to cater; since many algorithms suffer from problem of their local behavior.

Bug algorithms are fundamental and complete algorithms [9] with provable guarantees [10], since they let the robot to reach its destination if it lies in given space. In case the destination is not reachable, the robot has ability of terminating the assigned task. Each algorithm in Bug family carries its own termination property [11]. Bug algorithms do not suffer from local minima problem. In these algorithms the robot takes an action on the basis of current percepts of sensors without taking into account the previous path and actions. It is not a goal-oriented approach, as it does not consider goal's position and distance while avoiding an obstacle. It has two behaviors, "move to goal" and "obstacle avoidance". In obstacle avoidance behavior, it avoids an obstacle by just following the edges. It then changes the behavior to move to goal after avoiding the obstacle i.e. it restarts moving toward goal without considering any other parameter. Bug algorithms are divided into three types that differ from one another on the basis of their behavior of obstacle avoidance i.e. decision while encounter an obstacle.

### 2.1    Bug-1 algorithm

Bug-1 is the earliest obstacle avoidance algorithm, It is easy to tune and does not suffer by local minima however it takes the robot far away from the goal in some scenarios [9,12]. In this algorithm, the robot after detecting an obstacle starts following the edge of obstacle until it reaches to the point from where the robot started following the edge. It simultaneously calculates the distance from current position to destination and finally stores the point having minimum distance. This point, after one complete cycle of the robot is considered as leaving point. The robot restarts following the edge until it reaches to the calculated leaving point. After avoiding obstacle, robot computes new path from the leaving point $(x_1, y_1)$ to destination $(x_2, y_2)$ using straight line equation. The slop and y-intercept 'c' are given by (1) and (2) respectively.

$$m = \tan^{-1}\left(\frac{y_2 - y_1}{x_2 - x_1}\right) \tag{1}$$

$$c = y_1 - m \times x_1 \tag{2}$$

The robot follows that straight line until it reaches the destination or another obstacle is encountered. One of the common drawbacks of Bug-1 algorithm is, when the robot is following the edge of obstacle 1, it may collide with a neighboring obstacle 2 in case when the later is in very close proximity to the first obstacle or the gap between them is less than the width of the robot.

### 2.2    Bug-2 algorithm

Bug-2 algorithm is an improved version, which generates initial path from source to destination and stores slope $(m)$ of this path in its move to goal behavior. The behavior of the robot is changed to obstacle avoidance when an obstacle is encountered, where the robot starts following edge of the obstacle and continuously calculates slope of the line from its current position to the destination. When this slope becomes equal to slope of initial path (from source to destination), the behavior of the robot is changed to move to goal. Therefore, the robot follows single non-repeated path throughout its trajectory. Bug-2 algorithm is more efficient than Bug-1 algorithm as it allows the robot to reach the destination in less time following a short trajectory.

Both Bug1 and Bug2 algorithms demand minimum memory requirements. However, they do not have capability to make optimum use of sensors data for generation of shorter paths. An improved approach named as Dist-Bug algorithm addresses this problem [13].

### 2.3 Dist-Bug algorithm

Dist-Bug algorithm is final improved version of Bug algorithm series. It traverses comparatively shorter distance allowing the robot to reach destination in less time. This algorithm employs different obstacle avoidance behavior. When the robot encounters an obstacle in its path, it starts following the edge of the obstacle simultaneously calculating and storing the distance from its current and next position to destination. The leaving point, where it switches the behavior from obstacle avoidance to move to goal, is selected based on the condition that the distance of destination from its next position is greater than the corresponding distance from its current position (i.e. $(d_{next} > d_{current})$. The robot continues its obstacle avoidance behavior otherwise [14].

The objective of this research is to improve these algorithms by addressing their inherent problems and limitations. An algorithm having comparatively more intelligence and efficiency that can reach goal in comparatively less time by following smooth and short trajectory. The proposed, Intelligent Bug Algorithm (IBA) is attempted to achieve these objectives.

## 3 Proposed Algorithm : IBA

The detailed review of autonomous control strategies for mobile robot revealed that, in the category of Bug algorithms, Dist-Bug algorithm is most efficient as path cost is considered throughout the decision making process. However, it is not goal oriented and thus can take the robot far away from its goal position while avoiding obstacles. This is due to its leaving point decision during edge detection in obstacle avoidance behavior since the goal information is not taken into account. This gives the clue to improve Dist-Bug algorithm by devising an approach to make it goal oriented and to take time to destination into consideration. Based on this, the proposed IBA offers an intelligent control to navigate the robot in maze environment. Leaving point decision while detecting an obstacle in obstacle avoidance behavior of the robot in IBA is based on the goal position as well as the path cost. The later to reach destination takes time into account and makes the robot goal oriented. It improves the overall behavior of the robot and making it possible to achieve goal in comparatively less duration of time by following the short and smooth trajectory. Bug algorithms are unidirectional as they are able to take decision in one direction only. In contrast, bidirectional mechanism is introduced in IBA using the sensor's configuration on the robot and their Field Of View (FOV). The improved characteristics of IBA make it efficient to prove its convergence with relatively short and smooth trajectory.

The proposed IBA algorithm is based on two behaviors: *move to goal* and *obstacle avoidance*. Similar to Dist-Bug algorithm, the behaviors in IBA also depend on the present sensorial information of environment i.e. whether obstacles are sensed or not.

Initially, in *move to goal* behavior, a reference path is generated from source to goal position and the robot is forced to follow it until an obstacle is encountered or destination is reached. The behavior of the robot is changed to *obstacle avoidance* when an obstacle is sensed and the robot is commanded to follow the edges of the obstacle until leaving point is reached. In IBA, leaving point by taking the goal position into account is selected on the basis of free path toward the destination. The robot monitors the obstacles in the path towards destination while detecting edge in obstacle avoidance behavior. This condition, not introduced in Dist-Bug algorithm, offers goal orientation. The condition dictates that in IBA, the leaving point is not taken on the basis of minimum distance to destination. The obstacle-free path towards goal is also considered. This ensures that the robot does not have to wait for the point having minimum distance to goal. The robot changes its behavior to move to goal in order to generate new reference path, in case an obstacles-free path is sensed (just like a human as they follow the straight path after avoiding hurdles). The flowchart of IBA is shown in Fig. 1.

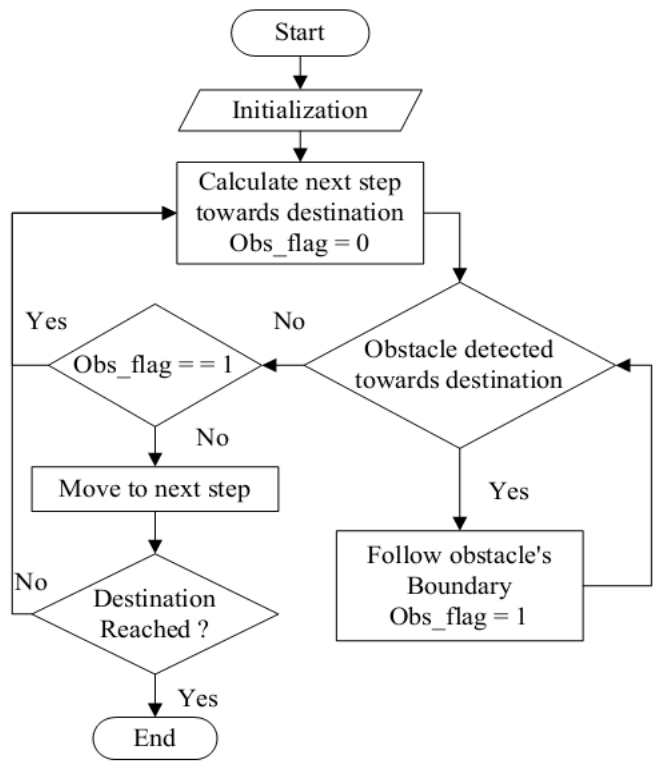

**Fig. 1.** Flowchart of IBA

## 4 Simulation Results

The effectiveness of the proposed algorithm (IBA) has been demonstrated using simulation results. The performance has been compared with other Bug algorithms reported in the literature. Consider an environment having a block shaped obstacle. The same environment is taken in all the cases to compare the performance of reported algorithms (Bug variants) and proposed algorithms. The designed simulation platform resulted in the robot trajectories corresponding to these algorithms. In case an obstacle is not sensed, the robot acts in a same manner in all the mentioned algorithms. Considering the scenario when no obstacle lies in the robot's path (Fig. 2a), it generates a path from source to destination and starts following it until it reaches to destination. The obstacle is now placed in the robot's path. Figure 2b illustrates the behavior of the robot in Bug-1 algorithm as it avoids both obstacles by edge detection and finds the leaving point finally reaching to destination successfully. Figure 2c shows the robot trajectory in Bug-2 algorithm where the robot is following the initial reference path by comparing the slop at each step while avoiding obstacles.

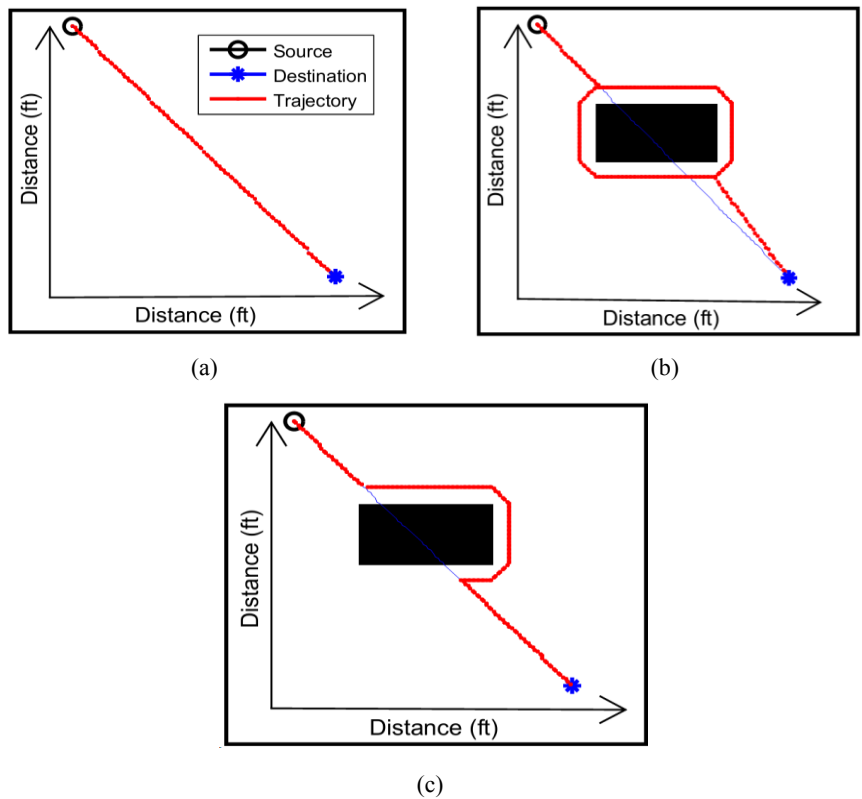

**Fig. 2.** Trajectories of Bug-1 and Bug-2 algorithms
(a) No obstacle  (b) Bug-1 algorithm (c) Bug-2 algorithm

Figure 3a shows the performance of the robot in Dist-Bug algorithm, where the robot is following the edge of obstacle until it reaches to the leaving point having minimum distance to destination. Simulation result of the proposed IBA is shown in Fig. 3b, the robot is following the edge until it finds the clear path towards the destination. Comparing the robot trajectories of Fig. 3a and 3b confirms that proposed algorithm improves the Dist-Bug algorithm since the path covered by the IBA is smaller and smoother than Dist-Bug algorithm.

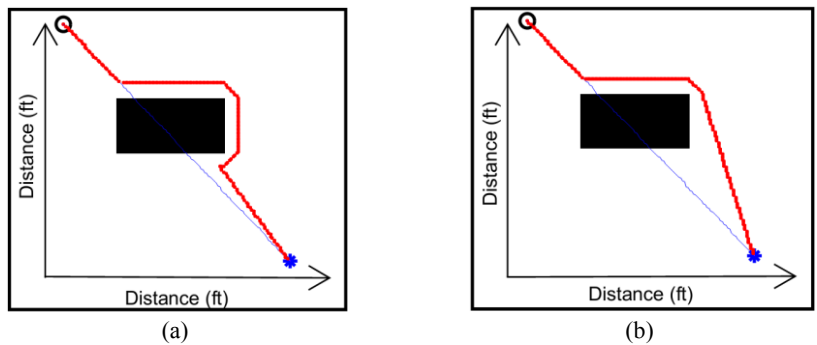

(a)          (b)

**Fig. 3.** Robot trajectories for performance comparison using
(a) Dist-Bug algorithm (b) IBA

The distance-time graph of above mentioned algorithms giving the path cost is illustrated in Fig. 4. The minimum distance from source to destination is 172 ft, which is covered in 120 sec when no obstacle lies in the path. However, this time increases by a factor dictated by algorithm efficiency in a path having obstacles. The most efficient variant of Bug algorithm i.e. Dist-Bug algorithm takes 179 sec where as the proposed IBA takes 162 sec to achieve the goal, which confirms the outstanding performance of IBA as compared to Dist-Bug algorithm.

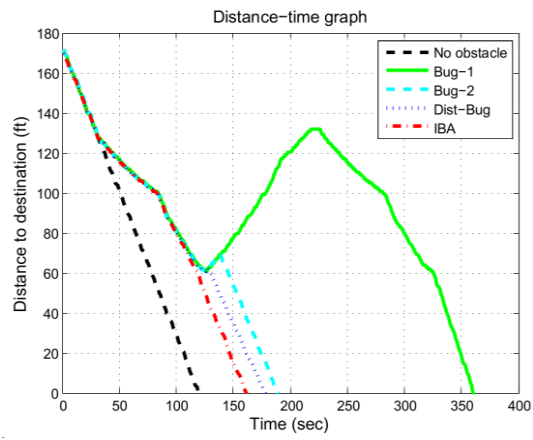

**Fig. 4.** Path cost v/s time for performance comparison

From the above comparison, it can be seen that the IBA takes least time and has a smoother trajectory than other bug algorithms. However, there are certain limitations of IBA that existing Bug algorithms also exhibit. These limitations are listed below. Some of the limitations can be overcome by fusing IBA with the bubble band technique proposed by Khatib and Quinlan in [15].

- IBA/Bug algorithms consider the robot as a circular point without taking its dimensions into account.
- The decisions are based on the basis of current percepts and therefore sensor noise may result in a wrong decision.
- The collision may be possible in the presence of an obstacle adjacent to the robot especially when avoiding the edges of the obstacle.

## 5   Conclusion

A new approach IBA is presented in this paper for autonomous navigation of mobile robots. The proposed algorithm IBA follows short and smooth trajectory and achieves the goal in less time as compared to reported Bug algorithms. IBA is a goal-oriented algorithm. The improved characteristics of IBA make it efficient to prove its convergence with relatively short and smoother trajectory in contrast with Dist-Bug algorithm. A bi-directional mechanism is introduced in IBA which can be achieved by using the sensor's configuration on the robot and their FOV. IBA, though exhibiting better performance than any algorithm in Bug family, needs to be further improved in an environment having U and H shaped obstacles. An on-going work to propose a more robust algorithm, Intelligent Follow the Gap Method (IFGM), considers time to destination into consideration. IFGM is planned to be intended for environments with symmetric and U/H-shaped obstacles by following the maximum gap among obstacles, thus ensuring more safer and shorter trajectory.